# Validating Pharmacogenomics Generative Artificial Intelligence Query Prompts Using Retrieval-Augmented Generation (RAG)

Ashley Rector
E: arector@mail.lipscomb.edu
P: 931-319-3616

Authors:
1. Ashley Rector, Pharmacy, Lipscomb University, Nashville, United States, PharmD, MHCI, MPharmSci
2. Keaton Minor, Helix Labs, Oklahoma City, United States, BSC
3. Kamden Minor, Helix Labs, Oklahoma City, United States, BSC
4. Jeff McCormack, Sherpa Rx, Franklin, United States, PhD, MS, HCLD (ABB), A(ACHI)
5. Beth Breeden, Pharmacy, Lipscomb University, Nashville, United States, DPh, MS
6. Ryan Nowers, Sherpa Rx, Franklin, United States, MAcc
7. Jay Dorris, Pharmacy, Lipscomb University, Nashville, United States, PharmD

**Introduction**

The integration of electronic health records (EHRs) and clinical decision support (CDS) systems in healthcare has paved the way for the practical application of pharmacogenomic data.[1] EHRs serve as comprehensive digital repositories of patient health information, which, when paired with CDS systems, enable evidence-based, personalized treatment recommendations.[2]

Pharmacogenomics is the study of how genetic variation influences drug response and is a rapidly advancing field, with the potential to revolutionize personalized medicine.[2,3] By tailoring drug selection and dosing based on genetic profiles, pharmacogenomics minimizes adverse drug reactions, reduces trial-and-error prescribing, and improves therapeutic outcomes.[2,3] The successful clinical implementation of pharmacogenomics is dependent on robust resources. One resource includes the Clinical Pharmacogenetics Implementation Consortium (CPIC) guidelines, which provide actionable recommendations for optimizing drug therapy based on genetic variants.[2] Additionally, Pharmacogenomics Knowledgebase (PharmGKB) is a curated resource compiling pharmacogenomic data, including drug-gene associations, clinical guidelines, and drug pathways, to support clinical decision-making.[4]

To address barriers in the adoption of pharmacogenomics, innovative technologies such as artificial intelligence (AI) and machine learning (ML) are being utilized.[6] Advanced models such as GPT-4, supported by RAG, enhance the accessibility and application of pharmacogenomic knowledge by embedding authoritative resources directly into their functionality.[3] Platforms such as Helix AI provide a robust backend for managing workflows and deploying AI applications in research and clinical settings.[7] In particular, one of Helix AI's specialized AI assistants is Sherpa Rx, which functions as a clinical guide that leverages large language models to provide accurate pharmacogenomics insights by directly integrating resources like CPIC guidelines into its context.[7,8] By embedding authoritative knowledge into AI-driven tools, these innovations can mitigate risks of inaccuracies and help provide reliable, actionable insights for healthcare providers.

Large language models (LLMs) have not been rigorously tested in pharmacogenomics, and their ability to accurately interpret and apply pharmacogenomic guidelines remains largely unknown. To address this gap, we constructed a benchmarking dataset designed to evaluate LLM performance on typical clinical pharmacogenomic queries. This dataset represents a novel standardized approach to assessing LLM capabilities in pharmacogenomics, offering an essential foundation for future evaluations.

This study's objective is to validate Sherpa Rx by evaluating the accuracy, completeness, and other key performance metrics of its responses to pharmacogenomic queries. The goal is to enhance clinical decision-making, foster knowledge sharing, and promote equitable access to pharmacogenomic insights.

**Materials and Methods**

For this study, Sherpa Rx was tailored to address specific PGx scenarios, focusing on queries related to CPIC guidelines. This method combines retrieval-based and generative techniques to provide contextually relevant, accurate responses beyond the capabilities of standalone generative models. The evaluation predominantly targeted healthcare providers, reflecting their primary role in utilizing pharmacogenomic insights for clinical decision-making. However, there was a subset of queries that targeted adults and pediatric patients, equipping them with clear insights into their genetic test results, potential health risks, and tailored treatment options.

*Knowledge Base Development*

The knowledge base (KB) incorporated CPIC guidelines and PharmGKB data. These resources were curated and structured to ensure comprehensive coverage of pharmacogenomic data, focusing on drug-gene interactions, dosing considerations, and therapeutic implications. In Phase 1, only the CPIC guidelines were integrated into the KB. In Phase 2, PharmGKB data were integrated into the KB with the CPIC guidelines to evaluate the impact of expanded resources on performance metrics. This content used the "text-embedding-3-small" model for generating embeddings, which are then stored and managed within a Pinecone vector database.

*Prompt Engineering and Guardrails*

Prompt engineering was used throughout Sherpa Rx to optimize the LLM's outputs for clinical relevance and structured responses. The system used prompts that define the LLM's role as "Pharmacogenomics Specialist" to ensure they are clinically oriented. For guardrails, Sherpa Rx relied on carefully designed prompts to maintain answer relevance and safety rather than implementing direct filtering or policy enforcement mechanisms. Sherpa Rx performed a straightforward cosine similarity search to find the most relevant documents. The temperature parameter was set to 0. The code did not implement a strict token limit, but the prompts themselves helped keep the overall token count minimized by prompting the models to develop concise responses.

*Dataset Selection*

A dataset comprising 260 pharmacogenomic queries was developed, with 10 queries representing each of the 26 CPIC guidelines available at the time of the study. The queries were created through a collaborative process: the principal investigator designed initial content based on published literature discussing pharmacogenomic drug-gene interactions, dosing considerations, and therapeutic recommendations.[1-3] A structured prompt was then used to guide ChatGPT-4omini in generating diverse, clear, and clinically meaningful variations of modeled pharmacogenomic queries tailored to specific CPIC guidelines. The CPIC guidelines served as the foundational framework throughout the process to ensure each query maintained clinical relevance and supported decision-making in practice. The dataset included queries corresponding to the following CPIC guidelines listed in Table 1:

| CPIC Guidelines |
| --- |
| CYP2B6 and efavirenz |
| CYP2C19 and clopidogrel |
| CYP2C19 and proton pump inhibitors |
| CYP2C19 and voriconazole |
| CYP2C9 and nonsteroidal anti-inflammatory drugs (NSAIDs) |
| CYP2C9 and HLA-B and phenytoin |
| CYP2C9, VKORC1, CYP4F2, and warfarin |
| CYP2D6 and atomoxetine |
| CYP2D6 and ondansetron and tropisetron |
| CYP2D6 and tamoxifen |
| CYP2D6, CYP2C19, and tricyclic antidepressants |
| Serotonin reuptake inhibitor antidepressants and CYP2D6, CYP2C19, CYP2B6, SLC6A4, and HTR2A |
| Opioids and CYP2D6, OPRM1, and COMT |
| CYP3A5 and tacrolimus |
| DPYD and fluoropyrimidines |
| G6PD deficiency |
| HLA-A, HLA-B, and carbamazepine and oxcarbazepine |
| HLA-B and abacavir |
| HLA-B and allopurinol |
| IFNL3 and peginterferon-alpha-based regimens |
| MT-RNR1 and aminoglycosides |
| RYR1, CACNA1S, and volatile anesthetic agents and succinylcholine |
| SLCO1B1, ABCG2, CYP2C9, and statins |
| TPMT, NUDT15, and thiopurines |
| UGT1A1 and atazanavir |

Table 1. CPIC guidelines incorporated.

*Evaluation Process*

To evaluate the AI assistant's performance and real-world applicability, a comprehensive assessment was conducted by a multidisciplinary team. This team included a Doctor of

Pharmacy (PharmD) candidate/dual master's candidate in healthcare informatics and pharmaceutical sciences, who led the evaluation process. The founder of Sherpa Rx (Ph.D. in Microbiology and Immunology) and two Cofounders of Helix Labs with engineering backgrounds designed and refined Sherpa Rx. Oversight was provided by two licensed pharmacists (PharmD/DPh) with expertise in clinical pharmacology, AI, and PGx.

*Performance Metrics and Scoring Criteria*

Responses generated that embedded CPIC guidelines alone were compared to those generated after incorporating PharmGKB data. All responses (N=260) in the first phase and a subset of query responses (N=20) in the second phase were assessed on a 5-point Likert scale for accuracy, relevance, completeness, and clarity, with recall also used for quantitative performance analysis. Precision and F1-score were calculated exclusively for 10 selected queries in Phase 1. The comparison was based on an annotation process that rated responses across the following criteria listed in Table 2:

| |
|---|
| Accuracy of Content: How correct and factually precise is the information provided? |
| 1: Completely inaccurate or wrong. |
| 2: Mostly inaccurate with some correct points. |
| 3: Somewhat accurate but contains significant errors. |
| 4: Mostly accurate with minor errors. |
| 5: Completely accurate and correct. |
| Relevance to Query: How well does the response address the specific question asked? |
| 1: Not related to the question at all. |
| 2: Slightly related but misses the main point. |
| 3: Moderately relevant, but not fully aligned with the question. |
| 4: Mostly relevant, covers most aspects of the question. |
| 5: Fully relevant and directly answers the question. |
| Completeness of Information: How thorough and detailed is the response? |
| 1: Very incomplete, missing most of the necessary information. |
| 2: Somewhat incomplete, lacking important details. |
| 3: Adequate but missing some information or details. |
| 4: Mostly complete, with minor gaps. |
| 5: Fully complete, providing all necessary details. |
| Clarity and Coherence: How clear and well-structured is the response? |
| 1: Confusing and poorly structured. |
| 2: Somewhat unclear and difficult to follow. |
| 3: Mostly clear but has some organizational issues. |
| 4: Clear and well-organized with minor issues. |
| 5: Very clear, well-organized, and easy to understand. |
| Recall: Assesses the model's ability to generate all relevant responses from the dataset |
| $$Recall = \frac{True\ Positives}{(True\ Positives + False\ Negatives)}$$ *True positives are considered as relevant points in the response that match or address the elements of the query, and false negatives are considered as relevant elements expected in the response but missing.* |
| Precision: Measures the proportion of relevant responses among the total responses generated by the model |
| $$Precision = \frac{True\ Positives}{(True\ Positives + False\ Positives)}$$ *False positives represent points in the response that are present but irrelevant or unrelated to the query.* |
| F1-Score: Provides a harmonic means of precision and recall, offering a single metric that balances both aspects. |
| $$F1-score = \frac{(Precision \times Recall)}{(Precision + Recall)} \times 2$$ |

Table 2. Assessment criteria

*Comparative Testing and Real-World Applicability*

A baseline comparison was also conducted using OpenAI's ChatGPT-4omini responses (N=20) to the same query subset assessed in Phase 2, with accuracy, relevance, completeness, clarity, and recall evaluated for these responses to further validate the improvements achieved with the expanded knowledge base. Additionally, a 20-question quiz addressing pharmacogenomic therapeutic scenarios was used to further assess the AI assistant's real-world applicability. Each question provided five answer choices and covered topics such as appropriate dosing and drug-gene interactions. A supplementary query layer was implemented to specifically target the drug and gene of interest to enhance Sherpa Rx's retrieval of pharmacogenomic data. Additionally, the total contextual input provided to the model was expanded to enhance accuracy, albeit with increased computational demands and extended processing time. Sherpa Rx's accuracy performance was benchmarked against ChatGPT-4omini, Claude 3.7 Sonnet, and Gemini 2.0 Flash.

*Statistical Analysis*

A one-tailed Wilcoxon signed-rank test ($\alpha = 0.05$) was conducted to compare accuracy scores across queries (N = 20) between Phase 1 and Phase 2. The test was repeated to compare accuracy between Phase 2 and ChatGPT-4o mini.

**Results**

*Context Management*

Figure 1 shows a multi-step process that outlines the functionality of the model, including query refinement, information retrieval, and context integration, to generate a comprehensive and relevant response.

Ensuring the AI assistant was generalizable to patients with a variety of backgrounds, including different ages, genders, races, and ethnicities, was a key priority. The goal was to ensure responses that encompassed friendliness, clarity, and understandability, while emphasizing the importance of shared decision making and consultation as needed. A detailed language and sentiment analysis of the results was outside the scope of this study, as the evaluation primarily relied on manual assessment and adjustments as needed.

# Sherpa Rx Query Processing Workflow

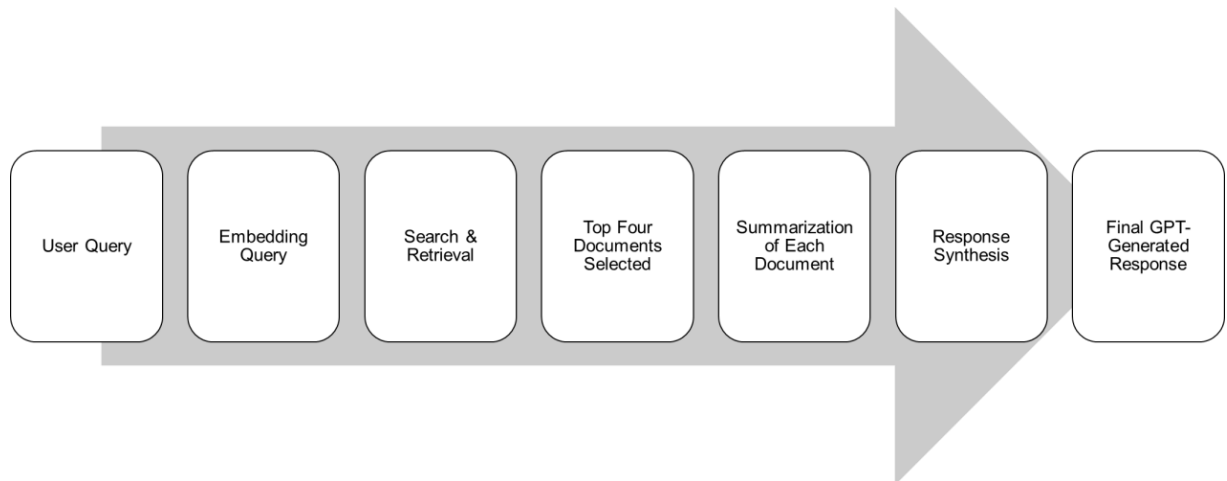

Figure 1. The Sherpa Rx tool processes user queries through a structured workflow to optimize response accuracy. First, the query is embedded, enabling efficient search and retrieval. Next, the system identifies and extracts the top four most relevant documents based on semantic similarity. Each document is then individually summarized to communicate key information. Finally, these summaries are synthesized into a comprehensive, GPT-generated response, ensuring a well-structured and contextually relevant answer.

*Provider Question:*
A 16-year-old Caucasian male with CF presents with compound heterozygosity for F508del and G551D CFTR mutations. How would you determine the appropriate dose of ivacaftor for this patient, considering their unique genetic profile?

*GPT-4omini Additional Prompt:*
N/A

*GPT-4omini Response:*

For a 16-year-old Caucasian male with cystic fibrosis (CF) compound heterozygous for F508del and G551D CFTR mutations, ivacaftor is indicated primarily for patients with a G551D mutation. The appropriate dosing for ivacaftor is typically 150 mg taken orally every 12 hours. However, for adolescents aged 6 years and older with CF and a G551D mutation, the dose remains the same, but it is crucial to consider:

1. **Weight-based dosing:** Ensure the patient's weight is considered, as dosing may need adjustment based on weight for those under 60 kg.
2. **Kidney and liver function:** Assess renal and hepatic function, as these may affect drug metabolism and dosing.
3. **Drug interactions:** Review potential drug interactions that may necessitate dose adjustments.

Consult the latest prescribing information for any updates or specific recommendations based on the patient's overall health and any other medications they might be taking.

Figure 2. ChatGPT-4omini's response in Phase 1 to a question from a healthcare provider.

*Provider Question:*
A 16-year-old Caucasian male with CF presents with compound heterozygosity for F508del and G551D CFTR mutations. How would you determine the appropriate dose of ivacaftor for this patient, considering their unique genetic profile?

*Sherpa Rx Additional Prompt:*
First layer: prompt = f"""Analyze the following pharmacogenomic document content from {source}, considering the user's query: "{query}"

Content:
{content}

Please provide a concise summary that addresses the query, keeping in mind:
1. If the query is directly related to pharmacogenomics, focus on relevant gene-drug interactions, genetic variants, and their clinical implications.
2. If the query is not directly related to pharmacogenomics, provide a brief overview of the document's content and explain how it might be relevant or not to the user's question.
3. If the query is a greeting or general question, acknowledge it and provide a brief introduction to the document's pharmacogenomic content.
4. Highlight any specific dosing recommendations, adverse effects, or efficacy information related to genetic factors, if relevant to the query.
5. If the query asks about a specific drug or gene not explicitly mentioned, look for information about related pharmacogenomic concepts or similar drug classes.
6. If the information seems incomplete or ambiguous, state this clearly in your summary.
7. Always try to relate the content back to the user's query, even if the connection is not immediately obvious."""

"content": "You are a pharmacogenomics specialist assisting users with various queries. Your task is to extract and summarize relevant information from pharmacogenomic documents, while always considering the user's specific query and intent."},

Second layer: "Based on the following summaries of pharmacogenomic documents and the user's query: "{user_input}", provide a comprehensive and relevant response.

   Summaries:
   {all_summaries}

   Please synthesize the information and provide a clear, structured answer, considering the following:
   1. Address the user's query directly, whether it's related to pharmacogenomics or not.
   2. If the query is a greeting or general question, provide a friendly response.
   3. For pharmacogenomics-related queries, highlight key gene-drug interactions, genetic variants, and their impact on drug efficacy or toxicity.
   4. If the query is not directly related to pharmacogenomics, explain how the available information might still be relevant or acknowledge that the information may not be directly applicable.
   5. Provide specific dosing recommendations or alternative treatments based on genetic factors, if applicable and relevant to the query.
   6. Highlight and point out any potential discrepancies between summaries.
   7. If there are ambiguities or incomplete information, clearly state this and provide possible interpretations based on current pharmacogenomic knowledge.
   8. Use a numbered or structured format for clarity and organize your response logically.
   9. If you need to use any knowledge outside of the provided summaries, clearly indicate this in your response.
   10. Always strive to make your response helpful and relevant to the user's specific query, even if it requires explaining why certain information may or may not be applicable."""
   try:
           "content": "You are an expert pharmacogenomics consultant. Your task is to provide a comprehensive, well-structured answer to a users query based on summaries from multiple pharmacogenomic documents. Use your expertise to make connections, infer information where appropriate, and highlight key pharmacogenomic concepts. "},

*Sherpa Rx Response:*
Based on the provided context, here are the steps to determine the appropriate dose of ivacaftor for a 16-year-old Caucasian male with CF who is compound heterozygous for F508del and G551D CFTR mutations: Genetic Testing Confirmation: Ensure that the genetic testing confirms the presence of both F508del and G551D CFTR mutations. Age Consideration: The patient is 16 years old, which falls within the age range (6 years and older) for which ivacaftor has been studied and recommended. Ivacaftor Indication: Ivacaftor is recommended for CF patients who have at least one G551D-CFTR variant. Since the

> patient has the G551D mutation, ivacaftor is indicated. Dosage Determination: Refer to the specific dosage guidelines for ivacaftor for patients aged 6 years and older. Typically, the dosage for patients aged 6 years and older is 150 mg every 12 hours. Consultation with CF Specialist: Ensure that the prescription and dosage are confirmed by a physician experienced in CF treatment, as recommended by the European Medicines Agency. Monitoring and Adjustments: Monitor the patient for efficacy and any potential side effects. Adjust the dosage if necessary, based on clinical response and any emerging safety data. Summary: Confirm genetic testing results. Patient age: 16 years old. Ivacaftor indication: Presence of G551D mutation. Recommended dosage: 150 mg every 12 hours. Consult with a CF specialist. Monitor and adjust as needed. This approach ensures that the patient's unique genetic profile and age are considered in determining the appropriate dose of ivacaftor.

Figure 3. Sherpa Rx's response in Phase 1 to a question from a healthcare provider.

*Performance Evaluation*

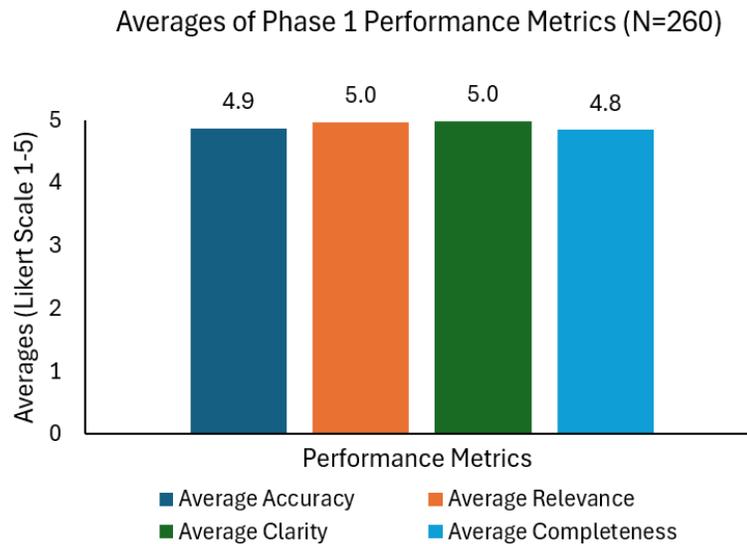

Figure 4. Averages of Phase 1 (N=260) Performance Metrics: Accuracy, Relevance, Clarity, and Completeness

Phase 1 (N=260) had overall averages of the following as shown in Figure 4: accuracy of 4.9, relevance of 5.0, clarity of 5.0, completeness of 4.8, and recall of 0.99.

For the subset analysis comparing Phase 1, Phase 2, and ChatGPT-4o mini (N=20), Phase 2 demonstrated improved accuracy (4.6 vs. 4.4) and completeness (5.0 vs. 4.8) over Phase 1, as shown in Figure 5. ChatGPT-4o mini achieved comparable scores for relevance (5.0) and clarity (4.9) but lower scores in accuracy (3.9) and completeness (4.2), as shown in Figure 5. Precision and F1-score were also calculated for 10 selected queries in Phase 1 relating to CPIC guideline, IFNL3 and peginterferon-alpha-based regimens, as shown in the Supplemental Index.

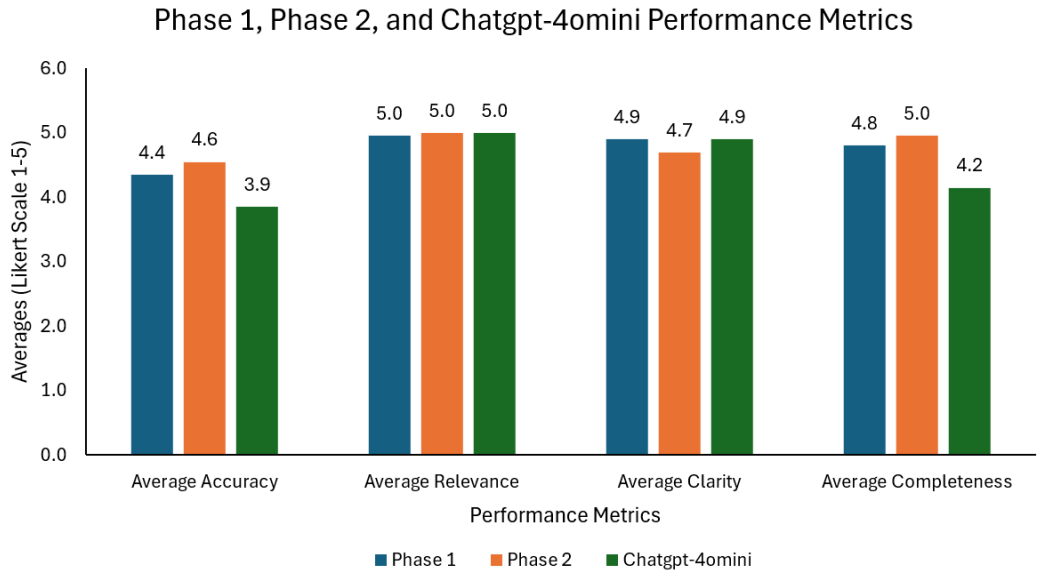

Figure 5. Averages of Phase 1, Phase 2, and ChatGPT-4omini Performance Metrics (N=20): Accuracy, Relevance, Clarity, and Completeness

| Comparison Group | Recall (Value: 0 to 1) |
|---|---|
| Phase 1 | 0.97 |
| Phase 2 | 0.99 |
| ChatGPT-4omini | 0.85 |

Table 3: Average Recall Comparison (N=20) between Phase 1, Phase 2, and ChatGPT-4omini

Table 3 shows that Phase 2 achieved the highest average recall (0.99) compared to Phase 1 (0.97) and ChatGPT-4omini (0.85), highlighting improved performance in the AI assistant's ability to generate all relevant responses from the dataset.

*Statistical Analysis*

While accuracy generally increased in Phase 2 compared to Phase 1, the Wilcoxon signed-rank test showed the difference was not statistically significant (W-value = 10.5, $p >$

0.05). However, Phase 2 demonstrated significantly higher accuracy compared to ChatGPT-4omini, as confirmed by the Wilcoxon signed-rank test (W-value = 18.0, p < 0.05).

*20-Question Quiz*

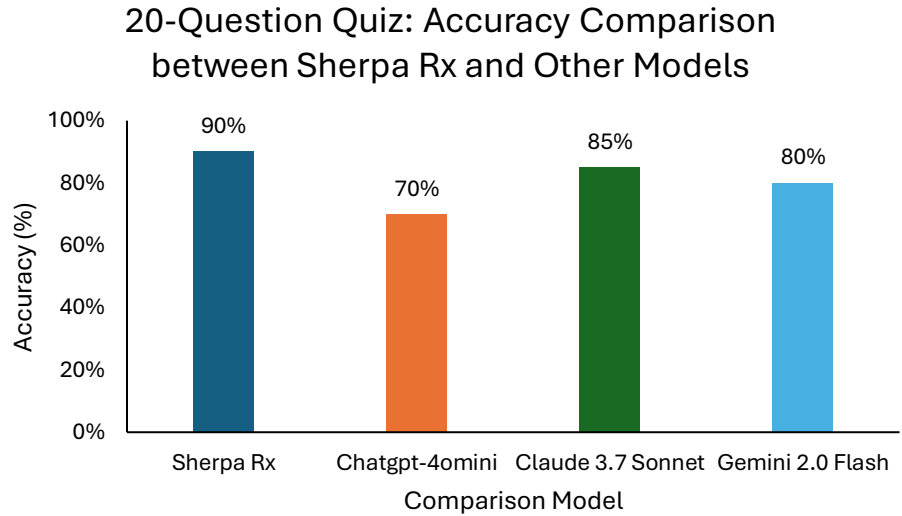

Figure 6: Accuracy comparison (N=20) between Sherpa Rx and other models on the real-world applicability quiz

Sherpa Rx's accuracy improved to 90%, outperforming ChatGPT-4omini, Claude 3.7 Sonnet, and Gemini 2.0 Flash, which scored 70%, 85%, and 80%, respectively. This enhancement of Sherpa Rx was achieved by adding a targeted query layer for the drug and gene of interest and expanding the model's context window, increasing accuracy at the cost of slightly slower performance (Phase 3).

**Discussion:**

The integration of a RAG approach with CPIC and PharmGKB data demonstrated a marked improvement in the AI assistant's accuracy and performance in handling pharmacogenomic queries. The AI assistant outperformed ChatGPT-4omini, particularly in accuracy and comprehensively addressing queries. This highlights the importance of incorporating authoritative, domain-specific resources into AI models to enhance their ability to deliver actionable information. The comparison between Phase 1 and 2 suggests that incorporating PharmGKB alongside CPIC guidelines provided value, but there are opportunities for further refining data integration and optimizing the use of domain-specific resources to enhance future performance.

     Despite these advancements, several limitations must be acknowledged. Ensuring accurate interpretation is critical, as errors in pharmacogenomic data integration could compromise patient safety. One limitation was the presence of multiple correct answer choices on one of the questions within the 20-question real-world applicability quiz, which, although accounted for in the analysis, introduces challenges in fully assessing response accuracy. Additionally, the small sample size (N=10) constrains the statistical power of performance metrics such as F1-score and precision, necessitating further validation with a larger dataset. The sample size in the subset analysis (N=20) is relatively small, and evaluating all queries (N=260), as in Phase 1, would increase the robustness. The accuracy and completeness of model predictions are also influenced by the complexity of integrating disparate data sources, particularly given patient-specific factors such as coexisting conditions, polypharmacy, and variability in renal and hepatic function. Furthermore, only one individual analyzed the query responses, which may introduce the possibility of errors or bias in the evaluation process. A multi-reviewer approach in future analyses could improve objectivity and consistency. Pharmacogenomic queries were partially generated using ChatGPT-4omini, which introduced minor typographical errors in approximately five queries, although this did not seem to affect the overall response classification. Future work should implement an expert panel to manually verify all generated queries to prevent such inconsistencies. A specific limitation identified in the query responses related to IFNL3 and peginterferon-alpha-based regimens was the difficulty in consistently differentiating between favorable and unfavorable genotypes for two distinct

polymorphisms (rs12979860 and rs8099917). Some responses incorrectly suggested that a genotype was favorable for both polymorphisms, despite differences in clinical implications. Future refinements should focus on enhancing data accuracy, consistency, and interoperability between CPIC, PharmGKB, and additional sources such as Pharmacogene Variation Consortium (PharmVar). Additionally, implementing a hybrid search framework that combines retrieval-augmented generation (RAG) with a fine-tuned pharmacokinetic model could improve response accuracy by leveraging both structured and unstructured data sources that address limitations related to accuracy and completeness.[8]

It is also important to emphasize that Sherpa Rx does not replace clinical judgment or the expertise of healthcare providers. Instead, it serves as a decision-support tool, designed to complement provider knowledge by delivering evidence-based, guideline-driven pharmacogenomic insights. Healthcare professionals remain responsible for interpreting AI-generated recommendations in the context of individual patient needs, ensuring that treatment decisions are personalized and clinically appropriate. Prospective validation in clinical settings will be critical to demonstrating scalability, reliability, and provider adoption before widespread implementation.

**Conclusion**:

This study highlights the transformative potential of generative AI in pharmacogenomics, emphasizing its ability to improve decision-making and patient care by providing accurate, personalized responses. The AI assistant demonstrated improvements in accuracy, relevancy, and real-world applicability, underscoring its potential to bridge knowledge gaps in precision medicine. With further analyses, refinement, and integration into healthcare systems, tools like this AI assistant can enhance provider productivity, support patient-centered care, and expand equitable access to pharmacogenomic data.

**Acknowledgments:**

N/A

**Competing Interests:**

No competing interests.

**Funding:**

No funding sources to disclose.

**References:**

1. Liu S, Wright AP, Patterson BL, et al. Using AI-generated suggestions from ChatGPT to optimize clinical decision support. *J Am Med Inform Assoc*. 2023;30(7):1237-1245. doi:10.1093/jamia/ocad072

2. Morris SA, Alsaidi AT, Verbyla A, et al. Cost effectiveness of pharmacogenetic testing for drugs with Clinical Pharmacogenetics Implementation Consortium (CPIC) guidelines: A systematic review. *Clin Pharmacol Ther*. 2022;112(6):1318-1328. doi:10.1002/cpt.2754

3. Murugan M, Yuan B, Venner E, et al. Empowering personalized pharmacogenomics with generative AI solutions. *J Am Med Inform Assoc*. 2024;31(6):1356-1366. doi:10.1093/jamia/ocae039

4. Whirl-Carrillo M, Huddart R, Gong L, et al. An evidence-based framework for evaluating pharmacogenomics knowledge for personalized medicine. *Clin Pharmacol Ther*. 2021;110(3):563-572. doi:10.1002/cpt.2350

5. Stryker C, Kavlakoglu K. What is artificial intelligence (AI)? IBM website. Updated

August 9, 2024. Accessed November 12, 2024. https://www.ibm.com/think/topics/artificial-intelligence#:~:text=Artificial%20intelligence%20(AI)%20is%20technology,and%20respond%20to%20human%20language.

6. Helix Private GenAIStack. HelixML,Inc website. Accessed November 12, 2024. https://tryhelix.ai/

7. Personalized & Precise Medicine Guided by AI. Sherpa Rx website. Accessed November 12, 2024. https://sherparx.com/

8. Gaedigk A, Casey ST, Whirl-Carrillo M, Miller NA, Klein TE. Pharmacogene Variation Consortium: A global resource and repository for pharmacogene variation. *Clin Pharmacol Ther*. 2021;110(3):542-545. doi:10.1002/cpt.2321

**Supplemental:**

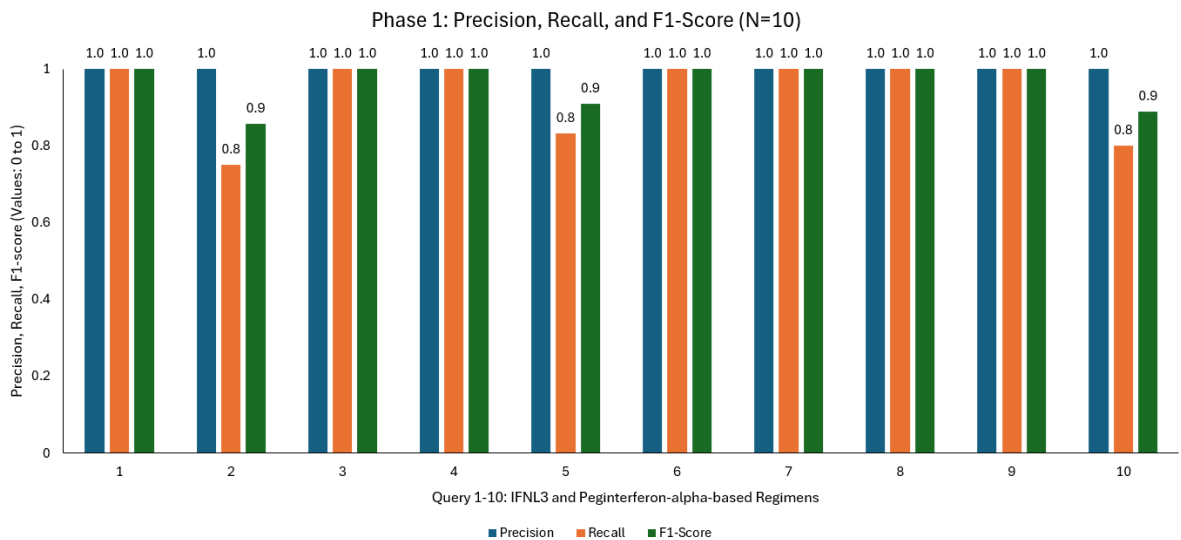

Figure 7: Precision, Recall, and F1-score for Phase 1-selected queries (N=10) of Sherpa Rx